\documentclass[runningheads]{llncs}

\usepackage{eccv}

\usepackage{eccvabbrv}

\usepackage{graphicx}
\usepackage{booktabs}

\usepackage[accsupp]{axessibility}  

\usepackage[pagebackref,breaklinks,colorlinks,citecolor=eccvblue]{hyperref}
\usepackage{hyperref}

\usepackage{orcidlink}

\usepackage{times}
\usepackage{epsfig}
\usepackage{graphicx}
\usepackage{amsmath}
\usepackage{amssymb}
\usepackage{todonotes}
\usepackage{multirow}
\usepackage{algorithm}
\usepackage{algpseudocode}
\usepackage{caption}
\usepackage{subcaption}
\usepackage{csquotes}
\usepackage{float}

\begin{document}

\title{Interactive Explainable Anomaly Detection for Industrial Settings} 


\author{Daniel Gramelt\inst{1,2} \and
Timon Höfer\inst{1} \and
Ute Schmid\inst{2}}

\authorrunning{D. Gramelt et al.}

\institute{
    Porsche Digital GmbH, Grönerstrasse 11/1, 71636 Ludwigsburg, Germany \inst{1} \\
    Otto-Friedrich-Universität Bamberg, Kapuzinerstraße 16, 96047 Bamberg, Germany \inst{2} 
}
\maketitle

\let\thefootnote\relax\footnotetext{The authors would like to thank Porsche AG for their support.}
\begin{abstract}
Being able to recognise defects in industrial objects is a key element of quality assurance in production lines. Our research focuses on visual anomaly detection in RGB images. Although Convolutional Neural Networks (CNNs) achieve high accuracies in this task, end users in industrial environments receive the model's decisions without additional explanations. Therefore, it is of interest to enrich the model's outputs with further explanations to increase confidence in the model and speed up anomaly detection. In our work, we focus on (1) CNN-based classification models and (2) the further development of a model-agnostic explanation algorithm for black-box classifiers. Additionally, (3) we demonstrate how we can establish an interactive interface that allows users to further correct the model's output. We present our NearCAIPI Interaction Framework, which improves AI through user interaction, and show how this approach increases the system's trustworthiness. We also illustrate how NearCAIPI can integrate human feedback into an interactive process chain. With this work, we plan to provide a new industry dataset for anomaly detection.
\end{abstract}

\section{Introduction}
\begin{figure}[ht] 
\centering
  \begin{minipage}[b]{0.24\linewidth}
    \includegraphics[width=\linewidth]{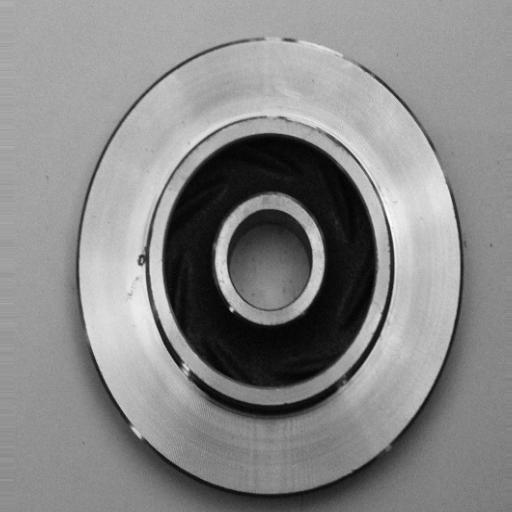}
    \caption*{(a)} 
  \end{minipage} 
  \begin{minipage}[b]{0.24\linewidth}
    \includegraphics[width=1\linewidth]{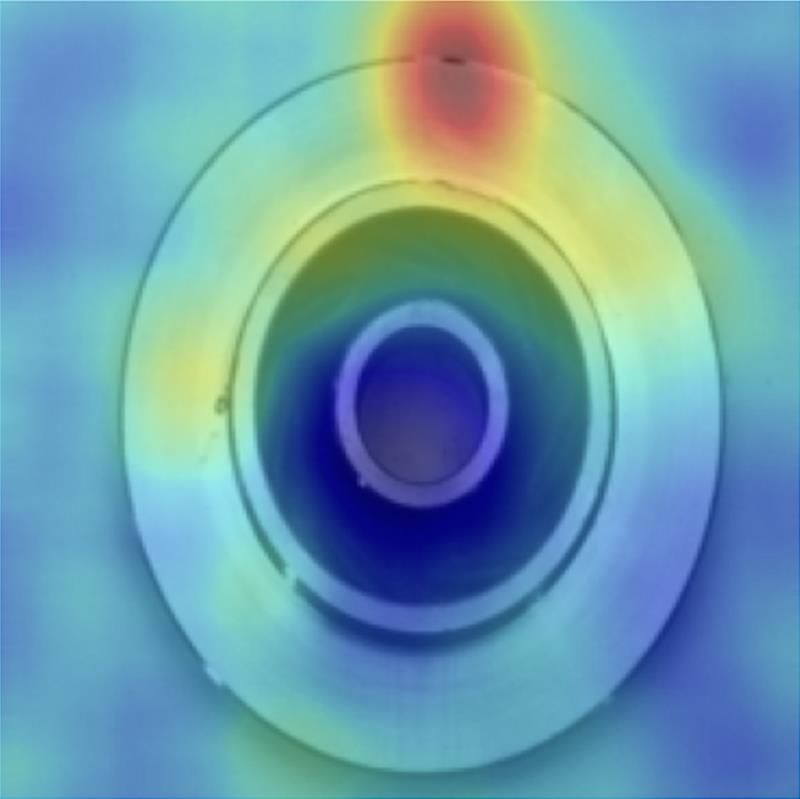} 
    \caption*{(b)} 
  \end{minipage} 
  \begin{minipage}[b]{0.24\linewidth}
    \includegraphics[width=\linewidth]{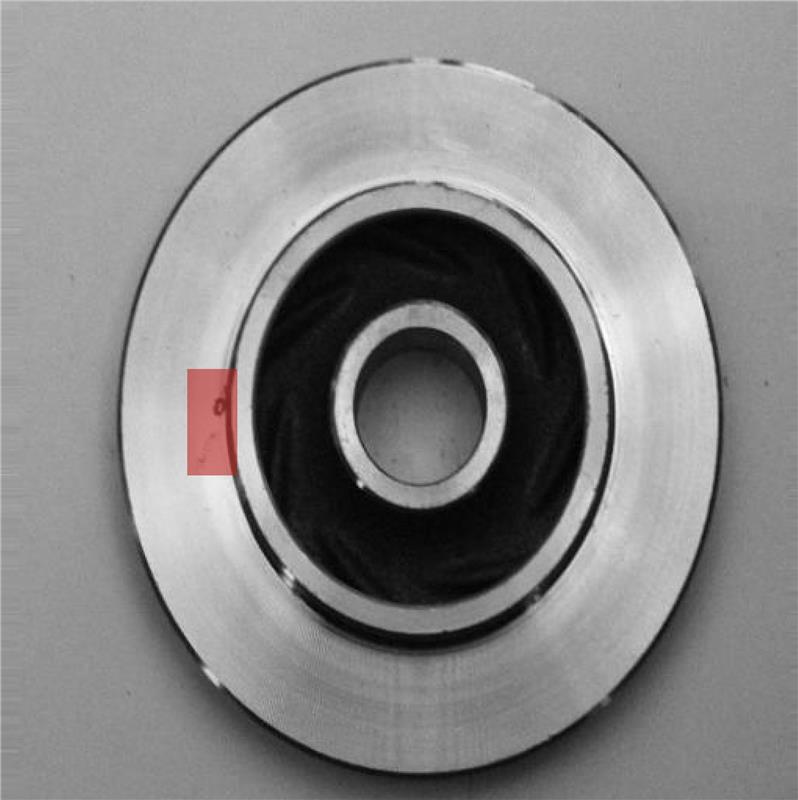} 
    \caption*{(c)} 
  \end{minipage}
  \begin{minipage}[b]{0.24\linewidth}
    \includegraphics[width=\linewidth]{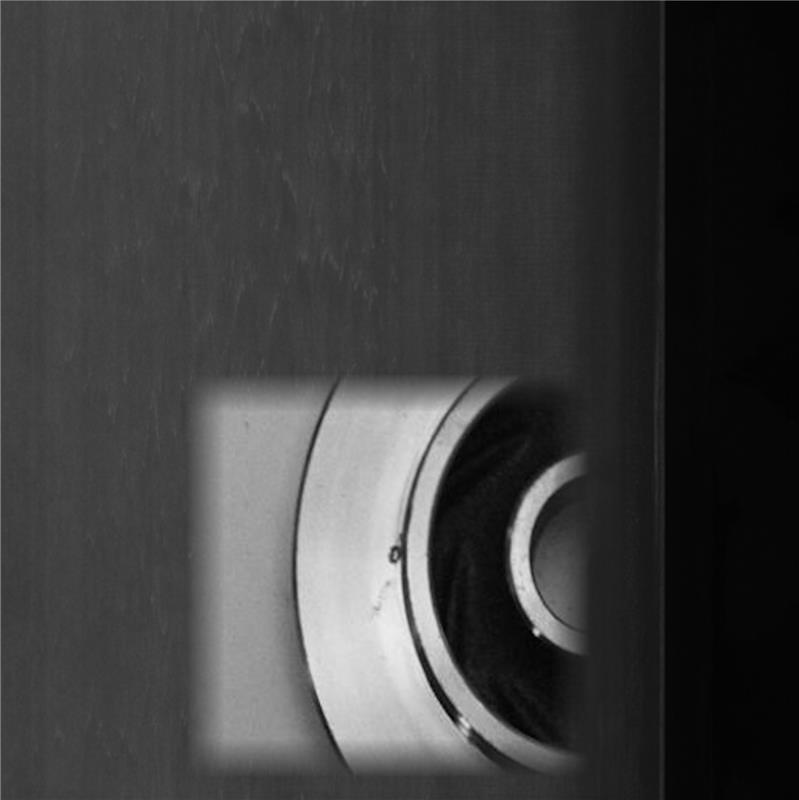} 
    \caption*{(d)} 
  \end{minipage} 
  \caption{An overview of the approach. (a) First, an image of a defective disc is presented. (b) Our classification model correctly predicts the model to be defect, but the explanation given is not correct. (c) A human then proceeds to inspect and correct the explanation by manually marking the area. (d) To enhance the performance of our model we generate further training data focusing on the area of the defect.} \label{InitialApproach} 
\end{figure}
 Anomaly detection, also known as outlier detection, plays a crucial role in our society. With the advancements in computer vision technology, numerous researchers are delving into the challenges of 2D anomaly detection, which encompasses detecting irregularities in both images \cite{reiss2021panda, pang2021explainable, roth2022towards, cohen2022transformaly} and videos \cite{georgescu2021anomaly, liu2021hybrid, chen2022comprehensive, liu2018future, wang2022video}. The goal of 2D anomaly detection is to pinpoint odd or unusual occurrences within visual data, such as photos and videos. This technique finds its use in multiple areas, including surveillance, security, healthcare imaging, and the inspection of industrial processes. Given the specific nature of the data, traditional statistical methods like isolation forest and k-NN, which are effective for structured data, are not directly applicable to visual data. Hence, methods based on deep learning are widely adopted for identifying anomalies in 2D data.

However, while it's been demonstrated that deep learning models are capable of spotting anomalies in visual content, the transparency behind how these models arrive at their conclusions is often lacking. Explainability refers to the model's ability to make its processes and decisions understandable to humans. In the context of 2D anomaly detection, an explainable model is expected to provide clear and reliable justifications for its judgments. Indeed, the issue of explainability is a significant barrier to the broader acceptance of data-driven approaches in the industrial sector \cite{langone2020interpretable, li2023survey, huang2022survey}. A typical use-case for anomaly detection in the industrial sector would be quality assurance in the production line. To speed up the finding process of the issue it is helpful to give end-users a reasoning for the classification decision, which can be done in the case of images by highlighting the pixels that are responsible for the anomaly detection. This also results in a higher trust in the model itself. Moreover, making AI systems explainable is not just an ethical imperative but also a legal one, particularly in sectors where human safety is at stake. Therefore, developing explainable models for 2D anomaly detection is crucial for supporting a wide range of human endeavours.
Common explainability methods for classification networks, such as RISE (Randomized Input Sampling for Explanation)\cite{petsiuk2018rise} focus on common object classification datasets such as COCO \cite{lin2014microsoft} or Pascal VOC \cite{everingham2010pascal} consisting of multiple different object categories. Anomaly detection in industrial settings only focuses on two classes: (1) the object is OK, or (2) the object is NOK because it has a scratch or some other anomaly. This introduces new problems, e.g. while blackening the pixel area of a dog in the image would result in not classifying the image as a dog anymore, in our scenario, by blackening the area of the scratches on a welding seam, we would still expect the model to classify the image as an anomaly. This also affects the usability of explainable methods such as RISE that were defined on the standard classification task.

Ultimately, the process should be designed to incorporate the human expert actively within the optimization process \cite{holzinger2016interactive}, enabling them to make interactive adjustments to both the explanations and the decisions. This approach aligns with interactive machine learning (ML) methods, which bear a resemblance to active learning \cite{settles2012active}. In active learning, the selection of instances and labels is facilitated through a collaborative effort between the algorithm and the user. Fulfilling these criteria ensures that the user retains comprehensive control over the entirety of the ML process by integrating the human expert into the workflow interactively, embodying the human-in-the-loop concept. In this work, one of our objectives is to analyze whether human feedback through explanation corrections enhances the model’s performance. Our primary contribution is aimed at enhancing the usability and performance of the CAIPI \cite{teso2019explanatory} algorithm, specifically within the field of industrial quality assurance. The CAIPI algorithm allows for the optimization of models by actively incorporating user feedback through the use of generated refutations that challenge predictions and explanations. We modified the CAIPI algorithm by using RISE \cite{petsiuk2018rise} instead of LIME \cite{ribeiro2016model}, introducing additional user feedback in case of correct prediction but wrong explanation \cite{slany2022caipi}, and incorporating near hits and misses \cite{herchenbach2022explaining} into the CAIPI algorithm.

Our contributions are as follows:
\begin{enumerate}
    \item We introduce an industrial dataset consisting of welding seams for an anomaly classification task.
    \item We introduce InvRISE, i.e. inverted RISE, a model agnostic explanation method specifically designed for anomaly classification.
    \item We install an extension of CAIPI \cite{teso2019explanatory} for human expert feedback by incorporating the idea of near hits and misses \cite{herchenbach2022explaining} which we name NearCAIPI.
\end{enumerate}

We will start by reviewing related work in section \ref{section2}, introduce our method in section \ref{section3} followed by the evaluation in section \ref{section4}.

\section{Related Work}\label{section2}
With the success of deep learning models based on convolutional neural networks (CNNs) for tasks such as image classification and object recognition, several classic backbone architectures such as VGG \cite{Simonyan15}, ResNet \cite{DBLP:journals/corr/HeZRS15} and ResNeXt \cite{xie2017aggregated} have become popular. However, they tend to be considered black box models because of the lack of transparency in the decision process.

In many scenarios, machine learning (ML) models need to present decisions in a transparent way, a requirement known as Explanatory AI (XAI). XAI methods can generally be divided into model-agnostic \cite{ribeiro2016model,petsiuk2018rise} and model-specific approaches \cite{ashish2017attention,herchenbach2022explaining}. Model-specific methods can produce impressive results but are limited to specific models. In contrast, model-agnostic methods offer broader applicability across different models. For example, LIME \cite{ribeiro2016model} (Local Interpretable Model-Agnostic Explanations) provides explanations for individual predictions. This distinction between local (individual predictions) \cite{petsiuk2018rise,scott2017shap} and global (overall model behaviour) \cite{herchenbach2022explaining,azzolin2023global} explanations is crucial to understanding and selecting appropriate XAI techniques. RISE \cite{petsiuk2018rise} (Randomized Input Sampling for Explanation), a model agnostic XAI method for estimating pixel saliency, has shown to outperform Lime in the image classification task and hence will be in focus for our work.

The idea of active learning \cite{monarch2021human,nissim2014novel,pfeuffer2023explanatory} is that a machine learning algorithm can reach higher levels of accuracy using fewer labeled training examples if it can select the data from which it learns. This approach is particularly well-suited for machine learning challenges where there is a plentiful supply of unlabeled data, but utilizing all data for training is unfeasible. In some cases, the reduced training set even improves performance \cite{schohn2000less}. 
Incorporating human feedback into the model development loop is another approach to enhance trust. In the work \cite{kulesza2015principles}, they describe the principles an interactive correction framework should follow. Namely: \emph{Be Actionable}, which makes the benefit for the user clear; \emph{Be Reversible}, because feedback can make a system worse than improving it; \emph{Always Honor User Feedback} for nudging the user to keep giving feedback; Lastly, \emph{Incremental Changes Matter} as showing the user the results and changes of their feedback will result in motivation and a better mental model of the user. Further considerations in regards to interactive learning can be found in \cite{herde2021survey,ouyang2022training, teso2023leveraging}.
CAIPI \cite{teso2019explanatory}, for instance, allows users to adjust faulty explanations, feeding corrected versions back into the dataset to improve accuracy. Hence, it falls under the category of interactive learning. To mitigate the learning of confound variables, \cite{schramowski2020making} utilize the CAIPI algorithm to overcome the "clever hans" behaviour. 
Furthermore, there are existing extensions to account for ethical correct behaviour (\cite{heidrich2023faircaipi}).

We are going to see how explainable methods have to be adapted for the task of anomaly classification by modifying the RISE explanations \cite{petsiuk2018rise}. Afterwards, we will utilize the concept of active learning in terms of a modified CAIPI \cite{teso2019explanatory} algorithm that builds upon the idea of near hits and misses \cite{herchenbach2022explaining}.

\section{Methodology}\label{section3}

In the following, we will focus on a setup where RGB images were taken of objects that were either okay or had defects, such as scratches. An initial overview of our approach can be found in Fig. \ref{Architecture}.

\subsection{Dataset}

While we also conduct experiments on the publicly available dataset \cite{CastingDataset}, we created a new dataset for anomaly classification of welding seams.

We introduce a dataset of self-made Metal Inert Gas (MIG) welding seams on aluminium plates. Here, a human expert produced 413 weldings where irregularities were placed on purpose. A welding seam is classified as correct if it is present, with a regular fish-scale, is fully bonded, has no cracks or pores and no interruption.

The final dataset consists of 413 images of 413 welding seams, where 139 are correct, 110 are welding plates without a welding seam and 164 are welding seams with irregularities. We classify welding seams as irregular if it has an irregular fish scale, if the welding seam contains black areas, if it has a binding error, if it has cracks or pores, or if it has an unfilled end crater.

All images were taken in the same setup. The camera was placed at a fixed distance with an angle of 90\textdegree. The resolution of the images is 1600 $\times$ 1200.
Each image is labeled by a human in one of the two categories: Welding regular (OK) and welding irregular (NOK). Due to strict limitations for a welding seam to be classified as OK we consider it as a challenging dataset for anomaly classification.

\begin{figure}[ht]
\centering
  \begin{minipage}[b]{0.30\linewidth}
    \includegraphics[width=0.99\linewidth]{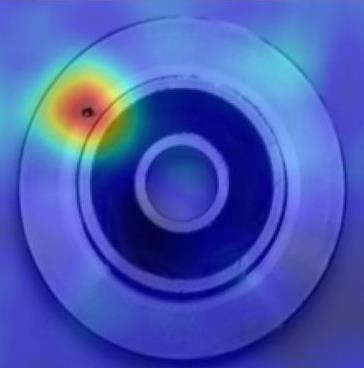}
    \caption*{(a)} 
   \end{minipage}\hspace{1cm}
  \begin{minipage}[b]{0.30\linewidth}
    \includegraphics[width=1\linewidth]{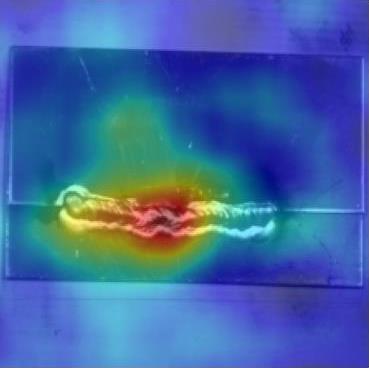} 
    \caption*{(b)} 
  \end{minipage} 
  \caption{Examples of correct explanations for images from the (a) casting product dataset and (b) our dataset consisting of welding seams.} \label{VisualExplanation} 
\end{figure}
\subsection{Classification}
Anomaly detection can be treated in multiple ways. Segmentation methods would be suited for detection of the anomaly area. The problem with segmentation models would be that they are restricted to finding the welding seams with information limited to the size. Irregularities, such as scratches, irregular fish scales would not be detected through this approach. 
Hence, we select the most general variant by treating the anomaly detection problem as a simple classification problem with the two classes: OK and NOK. The advantage of this approach is that the models we use can be lightweight which speeds inference and reduces the amount of compute recourses needed and are not limited to a specific kind of anomaly.

All ResNet \cite{DBLP:journals/corr/HeZRS15} models require an input size of 224  $\times$ 224. Therefore, all images were resized to a resolution of 224 $\times$ 224 before processing. 

\subsection{Explainability: InvRISE}
A well known method for generating local explanations of black box models is RISE \cite{petsiuk2018rise}. For the image classification task, RISE applies random masks to the image in order to find the relevant parts for the classification. Pixels within these masks are set to zero. Then the effect on the confidence for the predicted class is measured. If the confidence score goes down rapidly for a mask, then it means that the respective pixels were important for the class prediction.
In the case of predicting anomalies on objects, such as scratches, holes or any deformations, this approach could produce issues, e.g. as a blacked out area could be treated as a found anomaly. With this finding, we want to introduce a modified explanation method, which we call inverted RISE (\textbf{InvRISE}):

The first step is to sample k (we use k = 1000, higher values can also be considered) masks in the following way. We start with a quadratic matrix $L$ of dimension $l \in \mathbb{N}$ (we use l = 8), where we sample the matrix entries $l_{x,y}$ from a Bernoulli distribution with $p = 0.5$
\begin{align*}
    l_{x,y} \overset{d}{=} Ber(0.5), \; \;\;\;\; \forall \; (x,y) \in \{1, \dots, l\}^2.
\end{align*}
We then use linear up-sampling (duplicating existing pixels and averaging between neighbors) to map the small matrix $L \in \{0,1\}^{l,l}$ to the real valued matrix with similar dimension as our image $M \in \mathbb{R}^{224,224}$.

For a pixel $\lambda = (x,y)$ we approximate the probability of it being equal to 0 by the relative occurrence of this event on all masks $m$ following the distribution of $M$.
\begin{align}
    P[M_\lambda = 0 ] = \frac{\sum_m \mathbf{1}_{\{\lambda_m = 0\}}}{k}.
\end{align}
With this definition $P[M(\lambda) = 0]$ is the probability that the pixel $\lambda$ is not visible in the masks.

We define the inverted saliency map by
\begin{equation}\label{eq:InvRISE}
	S_{I,f}(\lambda)=\frac{1}{P[M_\lambda=0]}\sum_m (1 - f(I\odot m)) \cdot \overline{m}(\lambda) \cdot P[M=m].
\end{equation}

As a result, this is used to adjust the importance based on the likelihood of the pixel being not visible. In other words, the more often the pixel is not visible, the more informative the calculated value is. The result of the following formula is added up for each mask $m$ in the set of masks $M$. The image $I$ is combined with the mask $m$ to obtain a masked image. In the function $f(I \odot m)$ the masked image is used so that the network outputs a numeric confidence score, which describes how certain the image belongs to the target class. Subsequently, the confidence score is multiplied by $\overline{m}(\lambda)$. In this way, the calculated number is only taken into account for the importance if the pixel is not visible in the mask. This effect is ensured by $\overline{m}(\lambda)$ being $0$ if the pixel in $m$ is visible and $1$ otherwise. Lastly, the individual mask in the sum is weighted by $P[M = m]$ since this describes the probability that the mask $m$ is drawn from $M$. Therefore, masks that are applied multiple times are more influential when calculating the importance of pixels.

In detail, we make three changes in Equation \ref{eq:InvRISE}. First, we weigh the sum according to the probability that the pixel under consideration is obscured instead of visible. Second, we change the sum by $1-f(I\odot m)$, adding the confidence that the image is \emph{not} the target class. Finally, we negate $m(\lambda)$ from the original formula so that the value is only added up if the mask hides the pixel. With these three changes, we obtain an inversion of RISE, which considers a pixel to be relevant if the model does not predict the target class when masking the pixel. Example predictions can be found in Fig. \ref{VisualExplanation}.
\begin{algorithm}
\caption{Near CAIPI}
\begin{algorithmic}[1]
\State Select instance $x$ with the highest potential information gain from the unlabeled dataset $\mathcal{U}$
\State $\text{pred}_{x} \gets \text{AI}(x)$ 
\State $\text{exp}_{x} \gets \text{InvRISE}(x)$
\If{$\text{pred}_{x}$ and $\text{exp}_{x}$ are correct}
    \State $\mathcal{C}_x \gets \text{Refutations}(\text{exp}_x)$
\Else
    \State $\mathcal{C}_x \gets \text{Refutations}(\text{Correction}_{x})$
    \If{$\text{pred}_{x}$ is wrong}
        \State $x_{hit}, x_{miss} \gets \text{HitsAndMisses}(\mathcal{U}, x)$
        \State $\text{pred}_{hit} \gets \text{AI}(x_{hit})$
        \State $\text{pred}_{miss} \gets \text{AI}(x_{miss})$
        \State $\text{exp}_{hit} \gets \text{InvRISE}(x_{hit})$
        \State $\text{exp}_{miss} \gets \text{InvRISE}(x_{miss})$
        \If{$\text{pred}_{hit}$ and $\text{exp}_{hit}$ are correct}
            \State $\mathcal{C}_{hit} \gets \text{Refutations}(\text{exp}_{hit})$
        \Else
            \State $\mathcal{C}_{hit} \gets \text{Refutations}(\text{Correction}_{hit})$
        \EndIf
        \If{$\text{pred}_{miss}$ and $\text{exp}_{miss}$ are correct}
            \State $\mathcal{C}_{miss} \gets \text{Refutations}(\text{exp}_{miss})$
        \Else
            \State $\mathcal{C}_{miss} \gets \text{Refutations}(\text{Correction}_{miss})$
        \EndIf
        \State $\mathcal{T} \gets \mathcal{T} + [x_{hit},x_{miss}, \mathcal{C}_{hit}, \mathcal{C}_{miss}]$
    \EndIf
\EndIf
\State $\mathcal{T} \gets \mathcal{T} + [x,\mathcal{C}_x]$
\If{$N$ repetitions completed}
    \State Retrain AI with the extended training dataset $\mathcal{T}$
\EndIf
\end{algorithmic}
\end{algorithm}
\subsection{Human Interaction Pipeline}

\begin{figure}[ht] 
    \includegraphics[width=\linewidth]{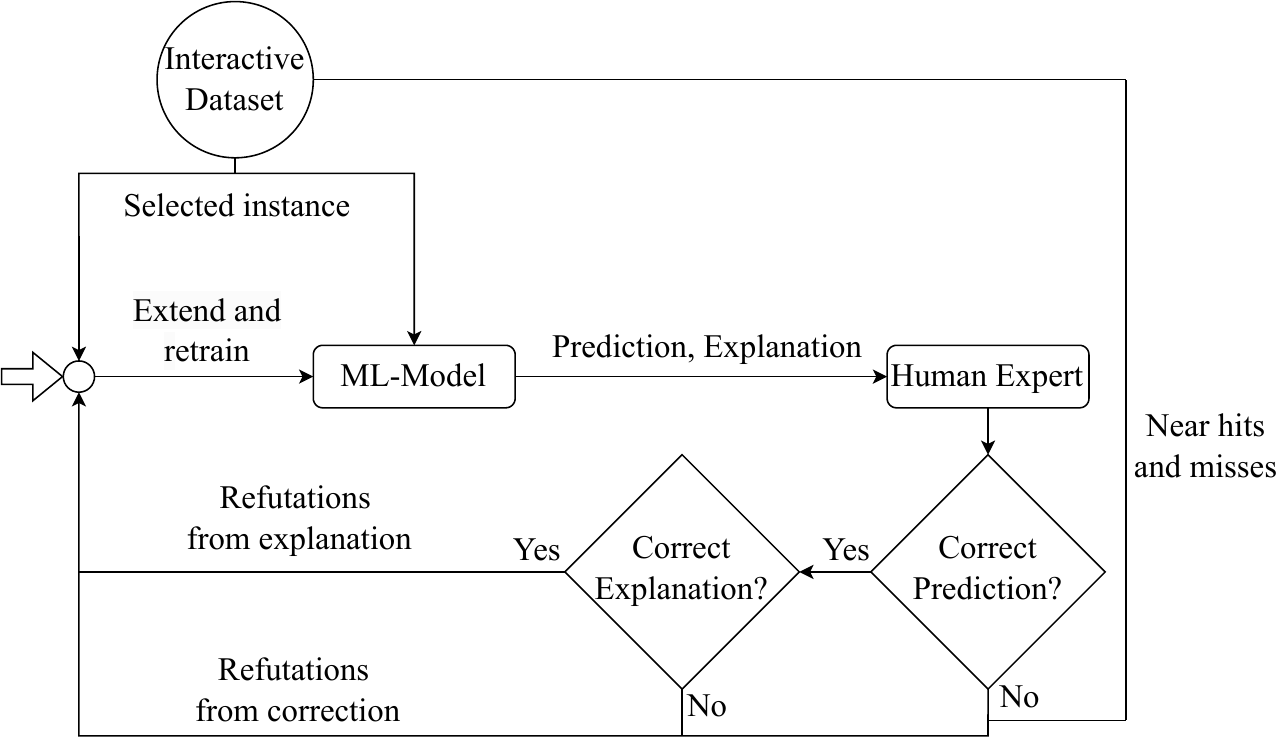}

    \caption{
    Illustration of how the human interaction pipeline works. First, an image with the highest potential for information gain is selected. For this image, the AI predicts the class and explains its decision to the human expert. We generate refutations depending on the expert's feedback and add the image and the refutations to the training dataset. If the prediction is wrong, we also expect feedback from the user regarding the nearest hit-and-miss of the image.} \label{Architecture} 
\end{figure}

Inspired by CAIPI \cite{teso2019explanatory} we want to introduce an interactive algorithm that brings a human into the loop. After training is complete, the human is provided with an interface where he is given examples of the model's output and is granted the option to correct predictions. The refinements from the human are then incorporated into the dataset to enable an iterative training procedure. Here, we assume to be given further data that is used only for the interactive component, which we call interactive dataset in the following. 

Following Fig. \ref{Architecture} we start with a classification model and select InvRISE as our explanation method. After training is complete we let our classification model run on the interactive dataset, where images with high uncertainty are presented to the user in an interface with the predicted class and the explanation given by InvRISE. The user can evaluate the prediction and explanation. Once the user has checked and, if necessary, corrected the explanation and prediction, new training data is generated out of this instance. After generating the new training data that is beneficial for the network, the images are added to the training dataset and removed from the interactive dataset. Using the extended training dataset, the model can be retrained or adapted to improve its performance. This loop is repeated until a specified number of iterations has been completed, no more images are available in the unknown dataset, or a specific performance of the network has been reached.

\begin{figure}[ht] 
  \begin{minipage}[b]{0.49\linewidth}
    \includegraphics[width=\linewidth]{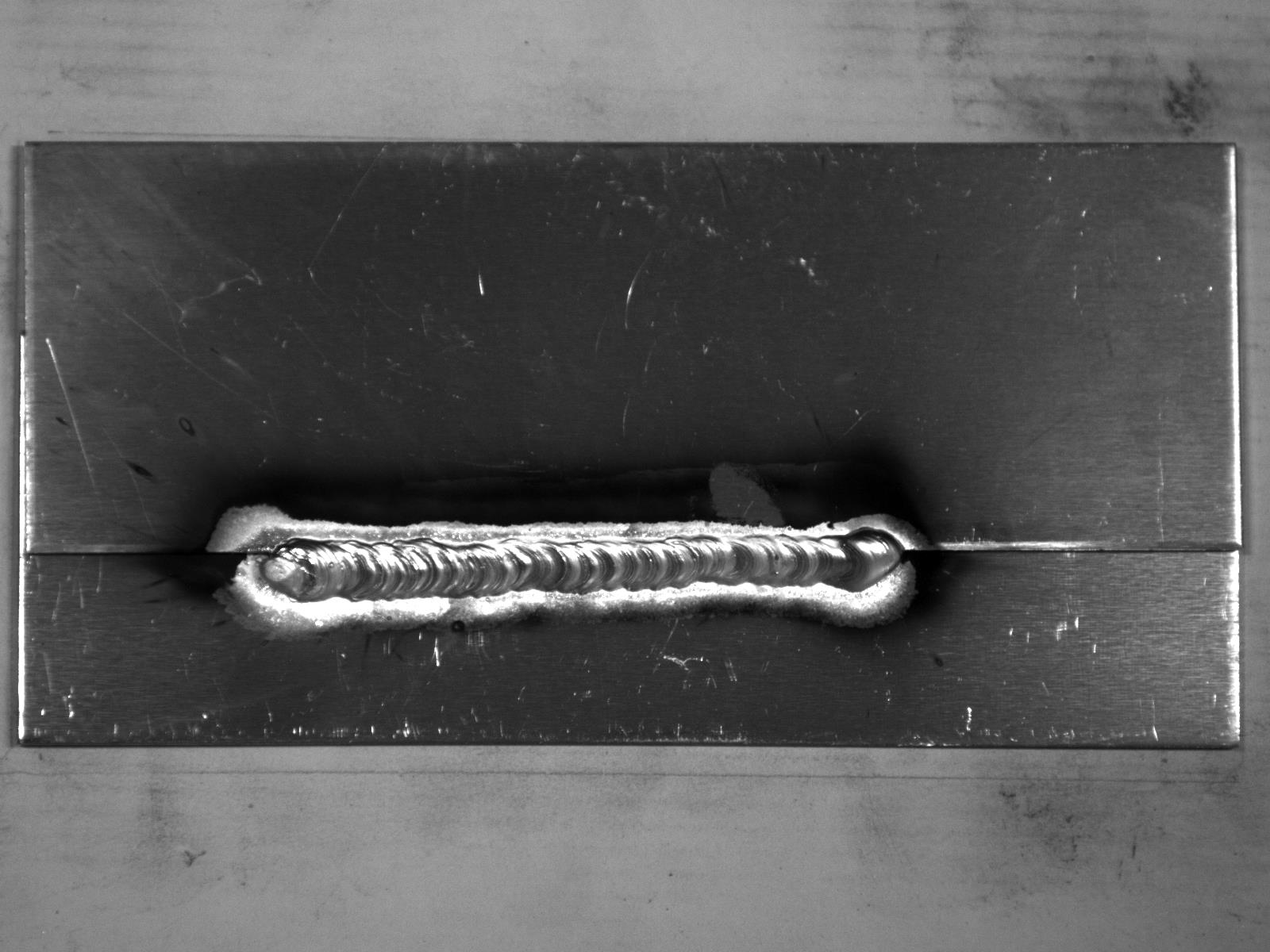}
    \caption*{(a) Input image: NOK} 
  \end{minipage} 
  \begin{minipage}[b]{0.49\linewidth}
    \includegraphics[width=1\linewidth]{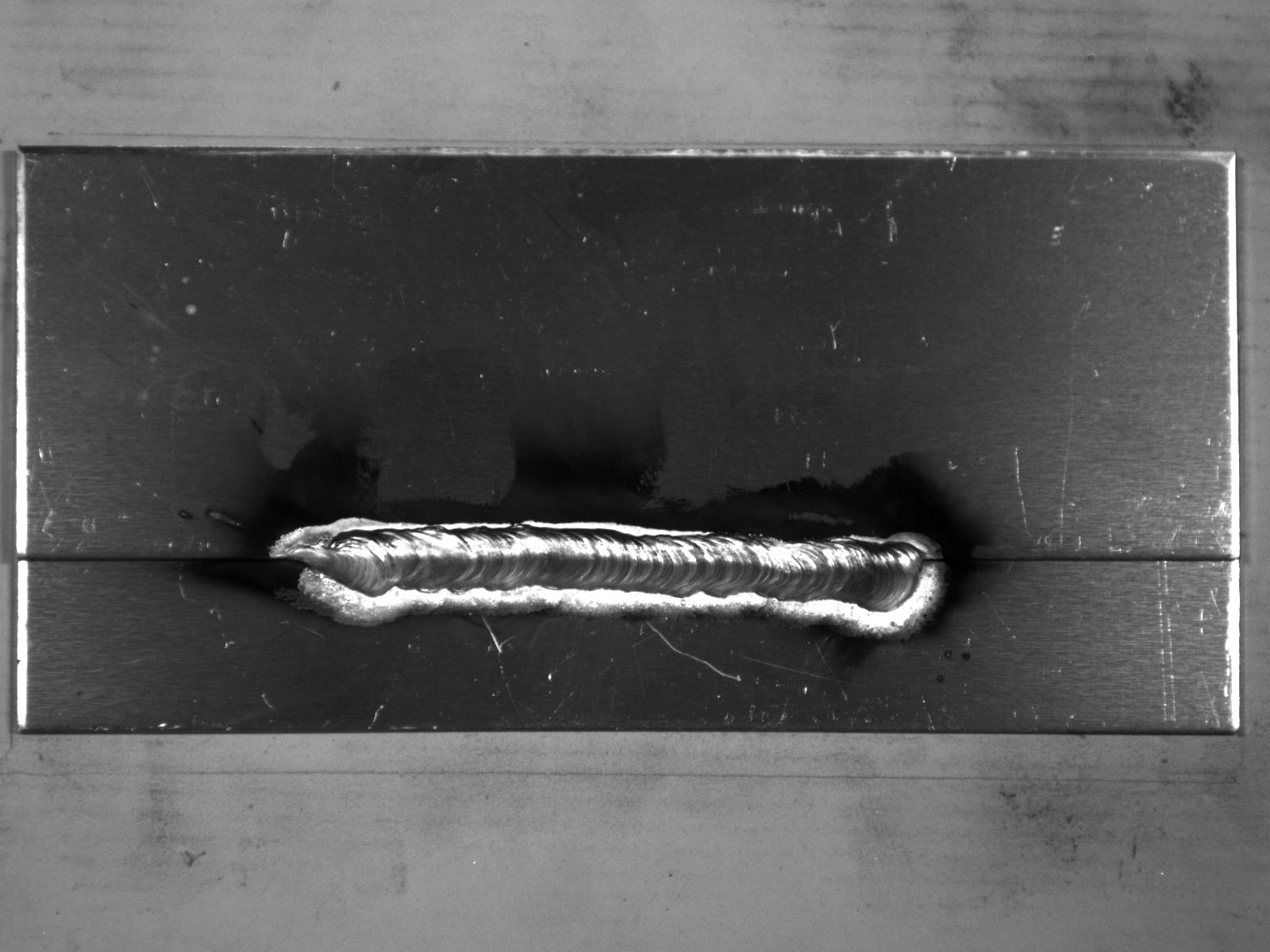} 
    \caption*{(b) Near hit: NOK} 
  \end{minipage} 
  \begin{minipage}[b]{0.49\linewidth}
    \includegraphics[width=\linewidth]{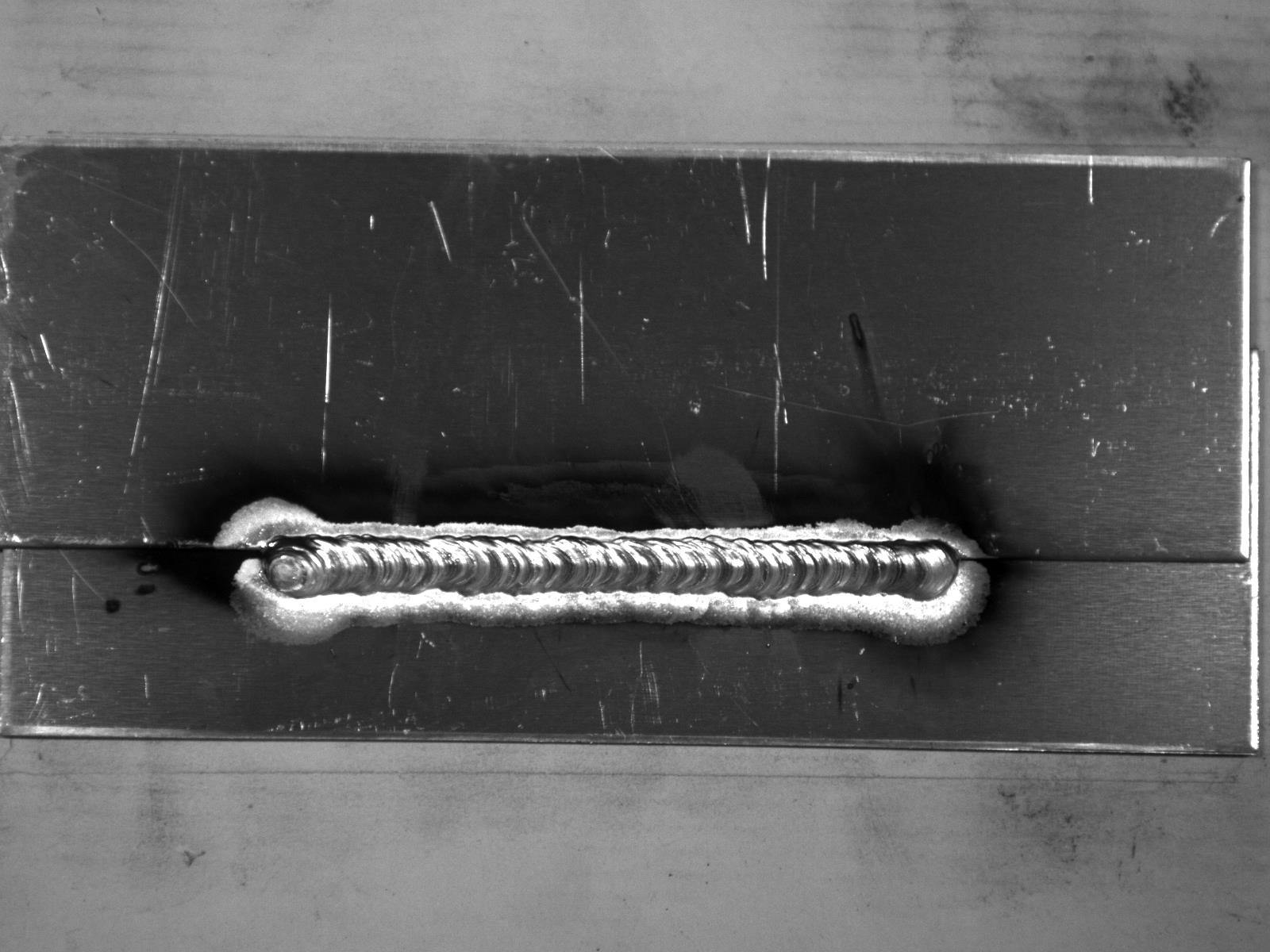} 
    \caption*{(c) Near miss: OK} 
  \end{minipage}
  \hfill
  \begin{minipage}[b]{0.49\linewidth}
    \includegraphics[width=\linewidth]{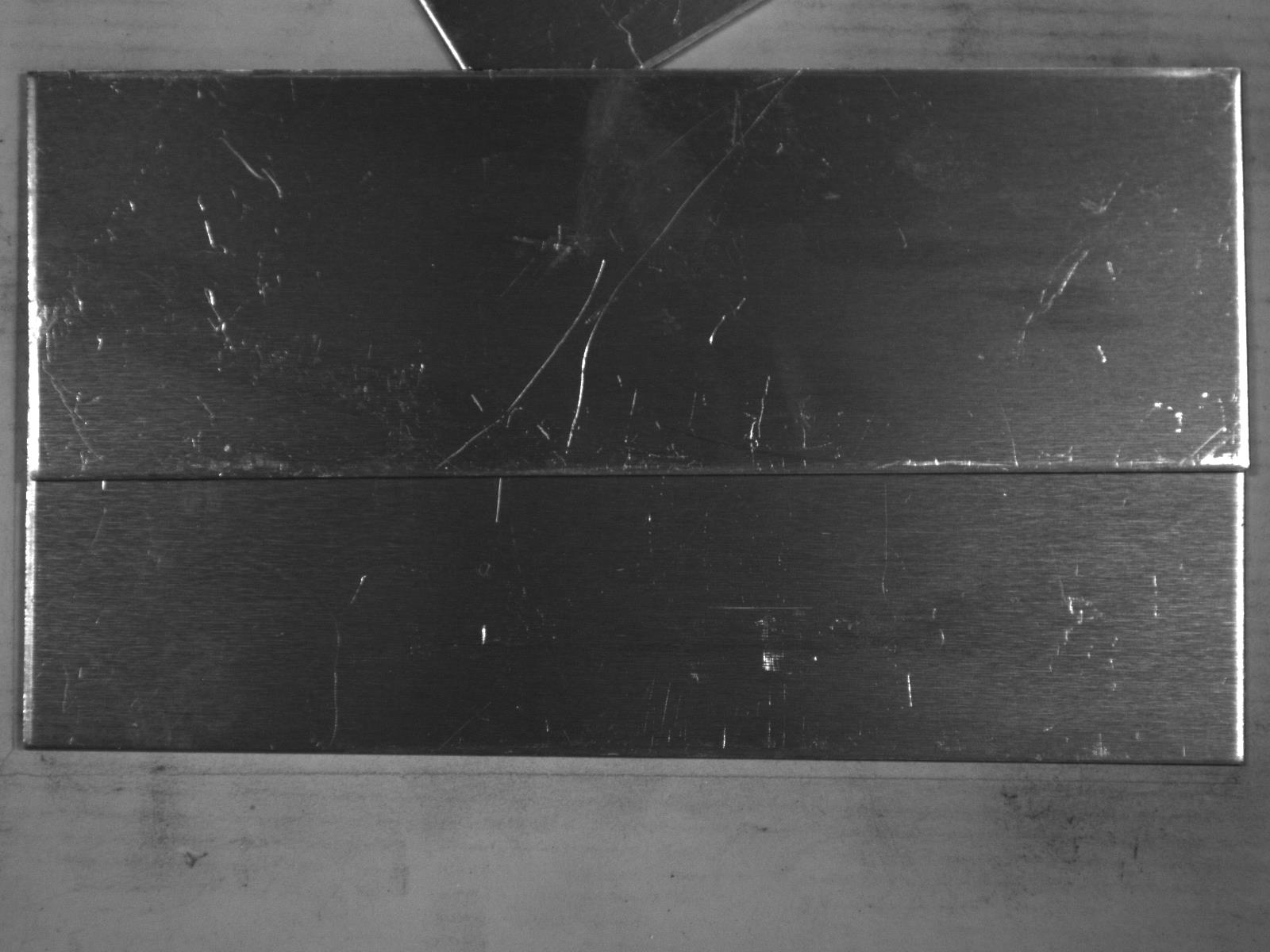} 
    \caption*{(d) Furthest hit: NOK} 
  \end{minipage} 
  \caption{An example of near hits and misses. (a) First, an image of the input image is presented. It consists of an irregular fish scale, and is therefore labeled as NOK. (b) We select the nearest image with the same label NOK; which is a welding seam that consists of an irregular welding seam and a possible binding error. (c) Additionally we show the nearest image with the label OK. (d) Lastly, we show the image, which is the furthest from our input image, which is a plate with no welding seam present.} \label{hitsandmisses} 
\end{figure}

We introduce an additional interaction step by using the feature embedding vectors of the images in order to find instances with similar features, which we call near hits/misses, inspired by \cite{herchenbach2022explaining}. Examples can be seen in Fig. \ref{hitsandmisses}. For instance, we generate a codebook for the images in the interactive dataset and calculate their feature embeddings $E_{\textit{feature}}$, which we define to be the representation in the second to last layer in the neural network. For a given image $x$ presented to the human, we calculate its feature embedding $E^x_{\textit{feature}}$ and select cosine similarity to be our distance metric to other embeddings $E^{x'}_{\textit{feature}}$ 
\begin{align}
     \textit{cosine}(E^x, E^{x'}) = \frac{E^x \cdot E^{x'}}{||E^x|| \; ||E^{x'}||}.
\end{align}
We then incorporate the idea of finding near hits, images with a similar embedding that have the same class, and near misses, images with a similar embedding that have a different class than the original image. The near hits and misses are then additionally presented to the human user to be inspected and potentially corrected, see Fig. \ref{hitsandmisses}.

Once the user has checked and, if necessary, corrected the explanation and prediction, new training data is generated out of this instance. After generating the new training data that is beneficial for the AI, the images are added to the training dataset and removed from the interactive dataset. Using the extended training dataset, the model can be retrained or adapted to improve its performance. Through this approach, we can detect and correct patterns of misclassified examples and thus more effectively improve the model's performance. 

\begin{table*}[ht!]
    \centering
\begin{tabular}{|c|c|c|c||c|c|}
    \hline
     & & \multicolumn{2}{|c||}{Welding} & \multicolumn{2}{|c|}{Casting}  \\
    \hline
    &Method& RISE & InvRISE & RISE & InvRISE  \\
    \hline
    \hline
    \multirow{5}{*}{AlexNet} & Model Acc. & \multicolumn{2}{|c||}{89\%} & \multicolumn{2}{|c|}{99\%} \\
    &Dice $\uparrow$ & 0.109 & 0.093 & 0.124 & 0.125 \\
    &Jaccard $\uparrow$& 0.075 & 0.061 & 0.075 & 0.075 \\
    &Hit Acc. $\uparrow$& 0.150 & 0.150 & 0.182 & 0.195 \\
    \hline
    \hline
    \multirow{5}{*}{VGG-16} & Model Acc. & \multicolumn{2}{|c||}{83\%} & \multicolumn{2}{|c|}{100\%} \\
    &Dice $\uparrow$& 0.076 & 0.078 & 0.245 & 0.240\\
    &Jaccard $\uparrow$& 0.046 & 0.053 & 0.153 & 0.150 \\
    &Hit Acc. $\uparrow$& 0.100 & 0.100 & 0.500 & 0.513 \\
    \hline
    \hline
    \multirow{5}{*}{ResNet-18} & Model Acc. & \multicolumn{2}{|c||}{86\%} & \multicolumn{2}{|c|}{100\%}\\
    &Dice $\uparrow$& 0.010 & 0.123 & 0.215 & 0.220 \\
    &Jaccard $\uparrow$& 0.061 & 0.073 & 0.131 & 0.135 \\
    &Hit Acc. $\uparrow$& 0.100 & 0.200 & 0.436 & 0.513\\
    \hline
    \hline
    \multirow{5}{*}{ResNeXt-50} & Model Acc. & \multicolumn{2}{|c||}{94\%} & \multicolumn{2}{|c|}{100\%} \\
    &Dice $\uparrow$& 0.100 & 0.089 & 0.182 & 0.185 \\
    &Jaccard $\uparrow$& 0.063 & 0.052 & 0.108 & 0.109 \\
    &Hit Acc. $\uparrow$& 0.130 & 0.217 & 0.308 & 0.282 \\
    \hline

\end{tabular}
    \caption{We compare different backbone architectures with additional explanation methods, here RISE and InvRISE. Higher values represent better performance for all metrics.}
    \label{tab:ExplanationExperiments}
\end{table*}

\section{Experiments}\label{section4}
For our experiments we conduct the evaluation on our own dataset consisting of 413 welding seams and a dataset of casting manufacturing product \cite{CastingDataset} consisting of 1,300 images. In addition, we use a deep metallic surface defect detection dataset \cite{BackgroundDataset} as background for the refutations.

\subsection{Experiments on Explainability}
Explanations are visualized via highlighted pixels, as seen in Fig. \ref{VisualExplanation}.
The explainability experiments were performed by training AlexNet, VGG-16, ResNet-18, and ResNeXt-50 using the welding and casting datasets. We had an expert annotate the irregular parts of the NOK images and compared the explanations with these annotations. In order to compare the generated saliency maps from RISE and InvRISE with a binary annotation, we also convert the saliency maps into a binary mask. To do this, we select the top 10\% of the pixels that are highlighted by InvRISE and mark them as important. Other pixels are therefore considered unimportant. With this method, we can compare the binary importance masks with the expert's binary annotated masks by using the Dice coefficient and the Jaccard metric. The Dice coefficient measures the similarity between two sets based on the ratio of the intersection to the total number of elements. In contrast, the Jaccard metric quantifies the similarity based on the ratio of the intersection to the union. In addition, we use a metric called hit accuracy, which tells us how often the most important pixel in the importance map was actually located in the region that the expert marked as important.
Conducted experiments can be found in Tab. \ref{tab:ExplanationExperiments}. Overall, we can see that in terms of accuracy and explanation, the welding seam dataset is more challenging. Older backbone architectures like AlexNet do not show a sufficient performance. We can see slight improvements from InvRISE over RISE in terms of the hit Accuracy. The Dice and Jaccard metrics show comparable results for both approaches. Overall, we see a slight improvement when using InvRISE and hence decided to use it as our explanation method for the interactive component.

\begin{table*}[h!]
    \centering
    \begin{tabular}{|l|lllllllllll|}
\hline
Retrainings                                       & \multicolumn{1}{l|}{} & \multicolumn{1}{c|}{1}      & \multicolumn{1}{c|}{2}      & \multicolumn{1}{c|}{3}      & \multicolumn{1}{c|}{4}      & \multicolumn{1}{c|}{5}      & \multicolumn{1}{c|}{6}      & \multicolumn{1}{c|}{7}      & \multicolumn{1}{c|}{8}      & \multicolumn{1}{c|}{9}      & \multicolumn{1}{c|}{10}     \\ \hline
\multirow{3}{*}{Random Add} & acc $\uparrow$                   & 0.65          & 0.70          & 0.74         & 0.78          & 0.78          & 0.83          & 0.81         & 0.82          & 0.79          & 0.82          \\
                           & f1 $\uparrow$                    & 0.67          & 0.64          & 0.67          & 0.66          & 0.72            & 0.78          & 0.74          & 0.77          & 0.72          & 0.72          \\
                          & mcc $\uparrow$                   & 0.39          & 0.38          & 0.46          & 0.52          & 0.54          & 0.64          & 0.59          & 0.63          & 0.56          & 0.64          \\ \hline
\multirow{3}{*}{AL}                               & acc $\uparrow$                   & 0.71          & \textbf{0.78} & \textbf{0.81}         & 0.81          & 0.86          & 0.84          & 0.85          & 0.88          & 0.87          & 0.89          \\
                                                  & f1 $\uparrow$                    & 0.65          & \textbf{0.74} & \textbf{0.76} & 0.76 & 0.81          & 0.78          & 0.79          & 0.83          & 0.84          & 0.86          \\
                                                  & mcc $\uparrow$                   & 0.41          & \textbf{0.55} & 0.60          & 0.61          & 0.71          & 0.66          & 0.67          & 0.74          & 0.73          & 0.77          \\ \hline
\multirow{3}{*}{Near AL}                          & acc $\uparrow$                   & \textbf{0.77} & 0.77          & \textbf{0.81} & 0.80          & 0.85          & 0.87          & 0.88          & 0.87          & 0.89          & 0.92          \\
                                                  & f1 $\uparrow$                    & \textbf{0.73} & 0.73          & \textbf{0.76}            & 0.76          & 0.79          & 0.82          & 0.83          & 0.84          & 0.86            & 0.90          \\
                                                  & mcc $\uparrow$                   & \textbf{0.53} & 0.54          & \textbf{0.61} & 0.59          & 0.67          & \textbf{0.72}          & 0.74          & 0.73          & 0.77          & 0.84          \\ \hline
\multirow{3}{*}{CAIPI}                            & acc $\uparrow$                   & 0.71          & 0.71          & 0.78          & \textbf{0.82} & \textbf{0.88} & 0.85          & 0.87          & 0.89          & \textbf{0.91} & 0.92          \\
                                                  & f1 $\uparrow$                    & 0.70          & 0.69          & 0.70          & \textbf{0.77}          & \textbf{0.84} & 0.79          & 0.83          & 0.84          & \textbf{0.89} & 0.90          \\
                                                  & mcc $\uparrow$                   & 0.46          & 0.46          & 0.54          & \textbf{0.63} & \textbf{0.76} & 0.69          & 0.72          & 0.78          & \textbf{0.82} & 0.84          \\ \hline
\multirow{3}{*}{Near CAIPI}                       & acc $\uparrow$                   & 0.74          & \textbf{0.78}          & 0.78          & 0.81          & 0.86          & \textbf{0.91} & \textbf{0.91} & \textbf{0.95} & 0.90          & \textbf{0.93} \\
                                                  & f1 $\uparrow$                    & 0.67          & 0.73          & 0.73          & 0.76          & 0.83          & \textbf{0.88} & \textbf{0.87} & \textbf{0.94} & 0.88         & \textbf{0.91} \\
                                                  & mcc $\uparrow$                   & 0.46          & 0.54         & 0.54          & 0.61          & 0.71          & 0.71 & \textbf{0.81} & \textbf{0.90} & 0.79         & \textbf{0.85} \\ \hline
\end{tabular}
    \caption{Comparison of the different (inter-)active learning approaches on the casting dataset. We compare random addition, active learning (AL), near active learning (Near AL), CAIPI, and NEAR CAIPI. In all metrics higher numbers represent better performance. We left 50 \% of the interactive data untouched for each of the different methods.}
    \label{tab:cipi_casting}
\end{table*}

\subsection{Experiments on Human Interaction}
The experiments were performed using ResNet-18 with the pre-trained IMAGENET1K\_V2 weights. We used early stopping with a patience of 10 epochs. As optimizer, we used SGD with a learning rate of 0.001 and momentum of 0.9.

The human interaction experiments were designed by splitting the entire dataset into four sub-datasets: training, validation, testing and interactive. All results presented were analyzed using the test dataset. The interactive dataset is the dataset from which images are transferred to the training dataset through user interactions. All images in the interactive data set were labelled and annotated in advance for automatic evaluation. It is, therefore, possible to simulate the user's decisions. However, since we cannot perfectly simulate whether a user accepts a given explanation, we always let the simulated user correct the images by the ground truth, guaranteeing good refutations.

We compare four methods to add new data from the interactive dataset to the training dataset. The first method is random addition, where random samples are selected and added to the training dataset. This corresponds to the normal procedure when a user labels new data. In the active learning (AL) approach, the AI predicts the entire unknown dataset and the instance with the lowest confidence is added to the training dataset. In the near active learning (near AL) approach, we add the nearest hit and miss of the wrong predicted image from the unknown dataset to the training dataset. CAIPI generates refutations and adds them to the training data set. We defined the refutations by zooming in or out on the anomaly and generating additional augmented versions of the images, as can be seen in Fig. \ref{InitialApproach}. Although no additional user interaction is required for the refutations, the user must evaluate the explanation and annotate the image if the prediction or explanation is wrong. This interaction is more time-consuming and intensive than active learning and random addition but takes the human into the loop. Finally, our proposed Near CAIPI approach additionally incorporates the idea of near hits and misses and generates refutations for them as well.
\begin{table*}[h!]
    \centering
\begin{tabular}{|l|llllllll|}
\hline
Retrainings                                       & \multicolumn{1}{l|}{} & \multicolumn{1}{c|}{1} & \multicolumn{1}{c|}{2} & \multicolumn{1}{c|}{3} & \multicolumn{1}{c|}{4} & \multicolumn{1}{c|}{5} & \multicolumn{1}{c|}{6} & \multicolumn{1}{c|}{7} \\ \hline
\multicolumn{1}{|c|}{\multirow{3}{*}{Random Add}} & acc $\uparrow$                   & 0.75                   & 0.75                   & 0.73                  & 0.73                  & 0.78                  & 0.78                  & 0.80                          \\
\multicolumn{1}{|c|}{}                            & f1 $\uparrow$                    & 0.58                & 0.64                 & 0.67                 & 0.67                 & 0.71                 & 0.71                 & 0.73                    \\
\multicolumn{1}{|c|}{}                            & mcc $\uparrow$                   & 0.41                 & 0.45                 & 0.48                 & 0.48                 & 0.55                 & 0.55                 & 0.59                  \\ \hline
\multirow{3}{*}{AL}                               & acc $\uparrow$                   & 0.68                  & 0.70                    & 0.80                    & 0.68                  & \textbf{0.85}          & 0.80                    & 0.75                      \\
                                                  & f1 $\uparrow$                    & 0.52                 & 0.50                    & 0.71                 & 0.65                 & \textbf{0.77}        & 0.71                 & 0.69                                \\
                                                  & mcc $\uparrow$                   & 0.27                 & 0.29                 & 0.57                  & 0.46                 & 0.57                  & 0.57                  & 0.52                \\ \hline
\multirow{3}{*}{Near AL}                          & acc $\uparrow$                   & \textbf{0.85}          & \textbf{0.90}           & \textbf{0.83}         & \textbf{0.80}           & 0.80                    & 0.85                   & 0.83                        \\
                                                  & f1 $\uparrow$                    & \textbf{0.73}        & \textbf{0.85}        & \textbf{0.76}        & \textbf{0.71}        & 0.75                   & 0.75                   & 0.67                       \\
                                                  & mcc $\uparrow$                   & \textbf{0.65}        & 0.77                 & \textbf{0.63}        & \textbf{0.57}         & 0.62                 & 0.65                 & 0.59             \\ \hline
\multirow{3}{*}{CAIPI}                            & acc $\uparrow$                   & 0.8                    & 0.725                  & \textbf{0.83}         & 0.75                   & 0.80                    & \textbf{0.88}         & 0.83                   \\
                                                  & f1 $\uparrow$                    & 0.71                 & 0.52                 & 0.74                 & 0.67                 & 0.73                 & \textbf{0.76}        & 0.76                      \\
                                                  & mcc $\uparrow$                   & 0.57                  & 0.34                  & 0.61                 & 0.48                 & 0.59                 & \textbf{0.72}        & 0.63                  \\ \hline
\multirow{3}{*}{Near CAIPI}                       & acc $\uparrow$                   & \textbf{0.85}          & \textbf{0.90}           & \textbf{0.83}         & 0.78                  & \textbf{0.85}          & 0.83                  & \textbf{0.88}    \\
                                                  & f1 $\uparrow$                    & \textbf{0.73}        & 0.82                 & 0.74                 & 0.69                 & 0.75                   & 0.74                 & \textbf{0.83}      \\
                                                  & mcc $\uparrow$                  & \textbf{0.65}        & \textbf{0.78}        & 0.61                 & 0.52                  & \textbf{0.65}        & 0.61                 & \textbf{0.74}                      \\ \hline
\end{tabular}
    \caption{Comparison of the different (inter-)active learning approaches on the welding seams dataset. We compare random addition, active learning (AL), near active learning (Near AL), CAIPI, and near CAIPI. In all metrics higher numbers represent better performance. We left 6 \% of the interactive data untouched for each of the different methods.}
    \label{tab:cipi_welding}
\end{table*}

We compare the performance of the models after each iteration, using a fixed number of interactions per iteration. We use accuracy, MCC (Matthews correlation coefficient), and F1-Score to compare performance. The MCC describes the quality of a classification model's predictions. A high MCC value implies that all classes were recognised with high accuracy, and therefore, the MCC is particularly meaningful for unbalanced datasets. The F1 measure is a harmonic mean between precision and recall and thus describes the balance between true positives, false positives, and false negatives.

Results can be seen in Tab. \ref{tab:cipi_casting}
for the Casting dataset, where we used 211 images as base training dataset and had 42 interactions per iteration.  
Here, we can see that all approaches are able to outperform the random addition baseline. For the first few iterations, the active learning approaches show the best results. With increasing iterations, the CAIPI algorithm outperforms the active learning, especially with the integration of near hits and misses. For both approaches, we find that introducing near hits and misses increases the performance.

For the Welding dataset, we used 135 initial training images of the welding dataset and having 27 interactions per iteration and the results are displayed in \ref{tab:cipi_welding}. Here, we can see that random addition and active learning are performing worse than the other methods. Again, adding near hits and misses increases the performance for both, active learning and CAIPI. Also CAIPI outperforms active learning with and without near hits and misses.

\section{Conclusion} \label{conclusion}

In this paper, we investigated the task of anomaly classification on industrial datasets consisting of objects that are defective, e.g. disks containing scratches or irregular welding seams. In particular, we explored the required vision modules for this problem, i.e. a CNN backbone for the classification extended with an additional explanation module, InvRISE. On top of that, we explored how humans can be incorporated via an interactive framework by extending CAIPI with the idea of near hits and misses. Both, the explanation module and the interactive framework increase trustworthiness for the human user and increase capabilities of the model. Experimental results show that the proposed framework is promising for industrial quality assurance.

%
%
\bibliographystyle{splncs04}
\bibliography{main}

\end{document}